# Comparative Performance Analysis of Image De-noising Techniques

Vivek Kumar, Pranay Yadav, Atul Samadhiya, Sandeep Jain, and Prayag Tiwari

*Abstract*—Noise is an important factor which when get added to an image reduces its quality and appearance. So in order to enhance the image qualities, it has to be removed with preserving the textural information and structural features of image. There are different types of noises exist who corrupt the images. Selection of the denoising algorithm is application dependent. Hence, it is necessary to have knowledge about the noise present in the image so as to select the appropriate denoising algorithm. Objective of this paper is to present brief account on types of noises, its types and different noise removal algorithms. In the first section types of noises on the basis of their additive and multiplicative nature are being discussed. In second section a precise classification and analysis of the different potential image denoising algorithm is presented. At the end of paper, a comparative study of all these algorithms in context of performance evaluation is done and concluded with several promising directions for future research work.

*Keywords*— Noise, Textural information, Image denoising algorithm, Performance evaluation.

## I. INTRODUCTION

A VERY large portion of digital image processing is deployed in image restoration. Image restoration is the removal or reduction of degradations which occurred while the image is being obtained [1]. Degradation in image comes from blurring as well as noise due to electronic and photometric sources. Blurring is a form of bandwidth reduction of the image caused by the imperfect image formation process such as relative motion between the camera and the original scene or by an optical system that is out of focus [2]. When aerial photographs are produced for remote sensing purposes, blurs are introduced by atmospheric turbulence, aberrations in the optical system and relative motion between camera and ground. In addition to these blurring effects, the recorded image is corrupted by noises too. A noise is introduced in the transmission medium due to a noisy channel, errors during the measurement process and during quantization of the data for digital storage. Each element in the imaging chain such as lenses, film, digitizer, etc. contributes to the degradation. Image denoising is often used in the field of photography or publishing where an image was somehow degraded but needs to be improved before it can be printed. Image denoising finds applications in fields such as astronomy where the resolution limitations are severe, in medical imaging where the physical requirements for high quality imaging are needed for analyzing images of unique events, and in forensic science where potentially useful photographic evidence is sometimes of extremely bad quality [2]. A two-dimensional digital image can be represented as a 2-dimensional array of data $s(x, y)$, where $(x, y)$ represent the pixel location. The pixel value corresponds to the brightness of the image at location $(x, y)$. Some of the most frequently used image types are binary, gray-scale and color images [3]. Binary images are the simplest type of images and can attain only two discrete values, black and white. Black is represented with the value '0' while white with '1'. Normally a binary image is generally created from a gray-scale image. A binary image finds applications in computer vision areas where the general shape or outline information of the image is needed. They are also referred to as 1 bit/pixel images.

Gray-scale images are known as monochrome or one-color images. They contain no color information. They represent the brightness of the image. An image containing 8 bits/pixel data means that it can have up to 256 (0-255) different brightness levels. A '0' represents black and '255' denotes white. In between values from 1 to 254 represent the different gray levels. As they contain the intensity information, they are also referred to as intensity images.

Color images are considered as three band monochrome images, where each band is of a different color. Each band provides the brightness information of the corresponding spectral band. Typical color images are red, green and blue images and are also referred to as RGB images. This is a 24 bits/pixel image.

## II. ADAPTIVE AND MULTIPLICATIVE NOISE

Basically the noises that corrupt the image are adaptive and multiplicative in nature. In this section all the major types of noises are being discussed.

*A. Gaussian Noise*

All Gaussian noise is evenly distributed over the signal [3].

Vivek Kumar is Lecturer and TPO with Laxmipati Group of Institutions, RGTU, Bhopal-462021, INDIA (Phone: +91 9098374992; e-mail: kunwarv4@gmail.com).
Pranay Yadav is with TIT College, RGTU, Bhopal-462021, INDIA (E-mail: pranaymedc@gmail.com).
Atul Samadhiya is with, IES college Bhopal-462021, INDIA (e-mail: aswoodstock40@gmail.com).
Sandeep Jain is with LIST college, RGTU, Bhopal-462021, INDIA (e-mail: jainsandeep10@hotmail.com).
Prayag Tiwari was with Millennium College, RGTU, Bhopal-462021, INDIA (E-mail: prayagtiwari2012@gmail.com).

This means that each pixel in the noisy image is the sum of the true pixel value and a random Gaussian distributed noise value. As the name indicates, this type of noise has a Gaussian distribution, which has a bell shaped probability distribution function given by,

$$F(g) = \frac{1}{\sqrt{2\pi\sigma^2}} e^{-(g-m)^2/2\sigma^2} \quad (1)$$

Where *g* represents the gray level, *m* is the mean or average of the function and σ is the standard deviation of the noise. Graphically, it is represented as shown in Fig.1. When introduced into an image, Gaussian noise with zero mean and variance as 0.05 would look as in Fig.1. Fig-2 illustrates the Gaussian noise with mean (variance) as 1.5 (10) over a base image with a constant pixel value of 100.

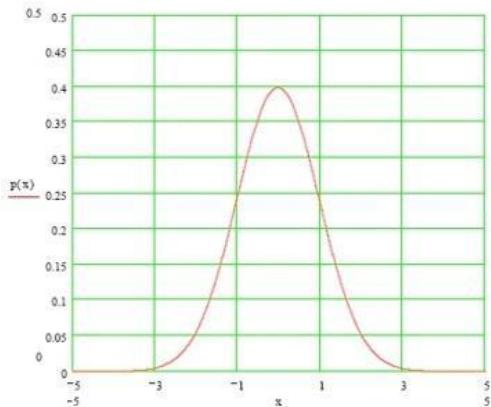

Fig. 1 Gaussian distribution

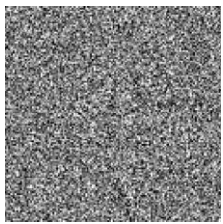 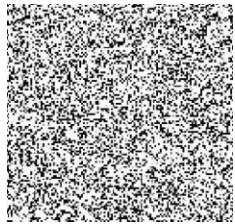

Fig. 2 (a) Gaussian noise  (b) Gaussian noise
(Mean=0, variance 0.05)  (Mean=1.5, variance 10)

### B. Salt and Pepper Noise

Salt and pepper noise [3] is an impulse type of noise, which is also referred to as intensity spikes. This is caused generally due to errors in data transmission. It has only two possible values *a* and *b*. The probability of each is typically less than 0.1. The corrupted pixels are set alternatively to the minimum or to the maximum value, giving the image a "salt and pepper" like appearance. Unaffected pixels remain unchanged. For an 8-bit image, the typical value for pepper noise is 0 and for salt noise 255. The salt and pepper noise is generally caused by malfunctioning of pixel elements in the camera sensors, faulty memory locations, or timing errors in the digitization process. The probability density function for this type of noise is shown in Fig-3. Salt and pepper noise with a variance of 0.05 is shown in Fig-4.

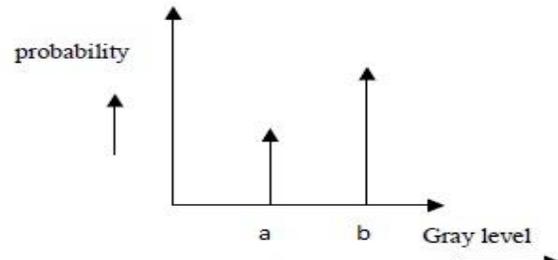

Fig. 3 Probability density function for salt and pepper noise

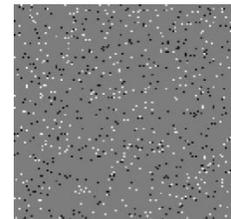

Fig. 4 Illustration of salt and pepper noise

### C. Speckle Noise

Speckle noise [4] is a multiplicative noise. This type of noise occurs in almost all coherent imaging systems such as laser, acoustics and SAR (Synthetic Aperture Radar) imagery. The source of this noise is attributed to random interference between the coherent returns. Fully developed speckle noise has the characteristic of multiplicative noise. Speckle noise follows a gamma distribution and is given as;

$$F(g) = \frac{g^{a-1}}{(\alpha-1)! \, a^\alpha} e^{-(g/a)} \quad \ldots (2)$$

where variance is $a^2\alpha$ and *g* is the gray level. On an image, speckle noise (with variance 0.05) looks as shown in Fig-6. The gamma distribution is given below in Fig-5.

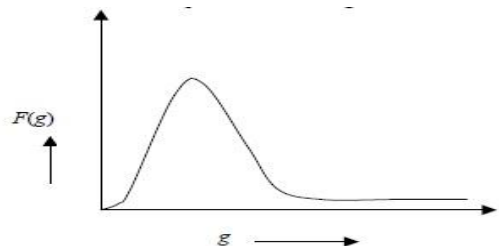

Fig. 5 Gamma distribution

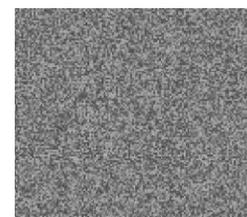

Fig. 6 Illustration of speckle noise

### D. Brownian Noise

Brownian noise [5] comes under the category of fractal or 1/*f* noises. The mathematical model for 1/*f* noise is fractional Brownian motion [Ma68]. Fractal Brownian motion is a non-stationary stochastic process that follows a normal distribution. Brownian noise is a special case of $1/f$ noise. It is obtained by integrating white noise. It can be graphically represented as shown in Fig-7. On an image, Brownian noise would look like Fig-8 which is developed from Fraclab [6].

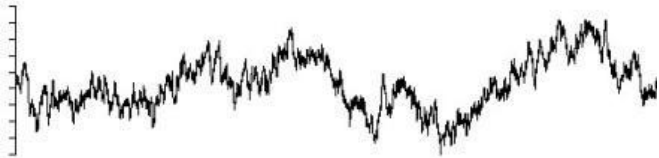

Fig. 7 Brownian noise distribution

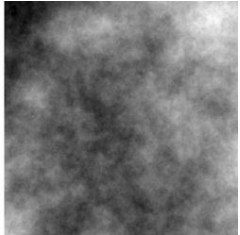

Fig. 8 Illustration of Brownian noise

### III. CLASSIFICATION OF DENOISING ALGORITHMS

On the basis of Fig.-1, it is obvious that there are two basic approaches of image denoising, spatial filtering methods and transform domain filtering methods.

*A. Spatial Filtering*

A traditional way to remove noise from image data is to employ spatial filters. Spatial filters can be further classified into non-linear and linear filters.

*1. Non-Linear Filters*

With non-linear filters, the noise is removed without any attempts to explicitly identify it. Spatial filters employ a low pass filtering on groups of pixels with the assumption that the noise occupies the higher region of frequency spectrum. Generally spatial filters remove noise to a reasonable extent but at the cost of blurring images which in turn makes the edges in pictures invisible. In recent years, a variety of nonlinear median type filters such as weighted median [8], rank conditioned rank selection [9], and relaxed median [10] have been developed to overcome this drawback.

*2. Linear Filters*

A mean filter is the optimal linear filter for Gaussian noise in the sense of mean square error. Linear filters too tend to blur sharp edges, destroy lines and other fine image details, and perform poorly in the presence of signal-dependent noise. The wiener filtering [11] method requires the information about the spectra of the noise and the original signal and it works well only if the underlying signal is smooth. Wiener method implements spatial smoothing and its model complexity control correspond to choosing the window size.

To overcome the weakness of the Wiener filtering, Donoho and Johnstone proposed the wavelet based denoising scheme in [12, 13].

*B. Transform Domain Filtering*

The transform domain filtering methods can be subdivided according to the choice of the basic functions. The basic functions can be further classified as data adaptive and non-adaptive. Non-adaptive transforms are discussed first since they are more popular.

*1. Spatial-Frequency Filtering*

Spatial-frequency filtering refers use of low pass filters using Fast Fourier Transform (FFT). In frequency smoothing methods [11] the removal of the noise is achieved by designing a frequency domain filter and adapting a cut-off frequency when the noise components are deco related from the useful signal in the frequency domain. These methods are time consuming and depend on the cut-off frequency and the filter function behavior. Furthermore, they may produce artificial frequencies in the processed image.

*2. Wavelet domain*

Filtering operations in the wavelet domain can be subdivided into linear and nonlinear methods.

*2.1 Linear Filters*

Linear filters such as Wiener filter in the wavelet domain yield optimal results when the signal corruption can be modeled as a Gaussian process and the accuracy criterion is the mean square error (MSE) [14], [15]. However, designing a filter based on this assumption frequently results in a filtered image that is more visually displeasing than the original noisy signal, even though the filtering operation successfully reduces the MSE. In [16] a wavelet-domain spatially adaptive FIR Wiener filtering for image denoising is proposed where wiener filtering is performed only within each scale and intrascale filtering is not allowed.

*2.2. Non-Linear Threshold Filtering*

The most investigated domain in denoising using Wavelet Transform is the non-linear coefficient thresholding based methods. The procedure exploits the property of the wavelet transform and the fact that the Wavelet Transform maps white noise in the signal domain to white noise in the transform domain. Thus, while signal energy becomes more concentrated into fewer coefficients in the transform domain, noise energy does not. It is this important principle that enables the separation of signal from noise. The procedure in which small coefficients are removed while others are left untouched is called hard thresholding [7]. But the method generates spurious blips, better known as artifacts, in the images as a result of unsuccessful attempts of removing moderately large noise coefficients. To overcome the demerits of hard thresholding, wavelet transforms using soft thresholding was also introduced in [7]. In this scheme, coefficients above the threshold are shrunk by the absolute value of the threshold itself. Similar to soft thresholding, other techniques of applying thresholds are semi-soft thresholding and Garrote thresholding. Most of the wavelet shrinkage literature is based on methods for choosing the optimal threshold which can be adaptive or non-adaptive to the image.

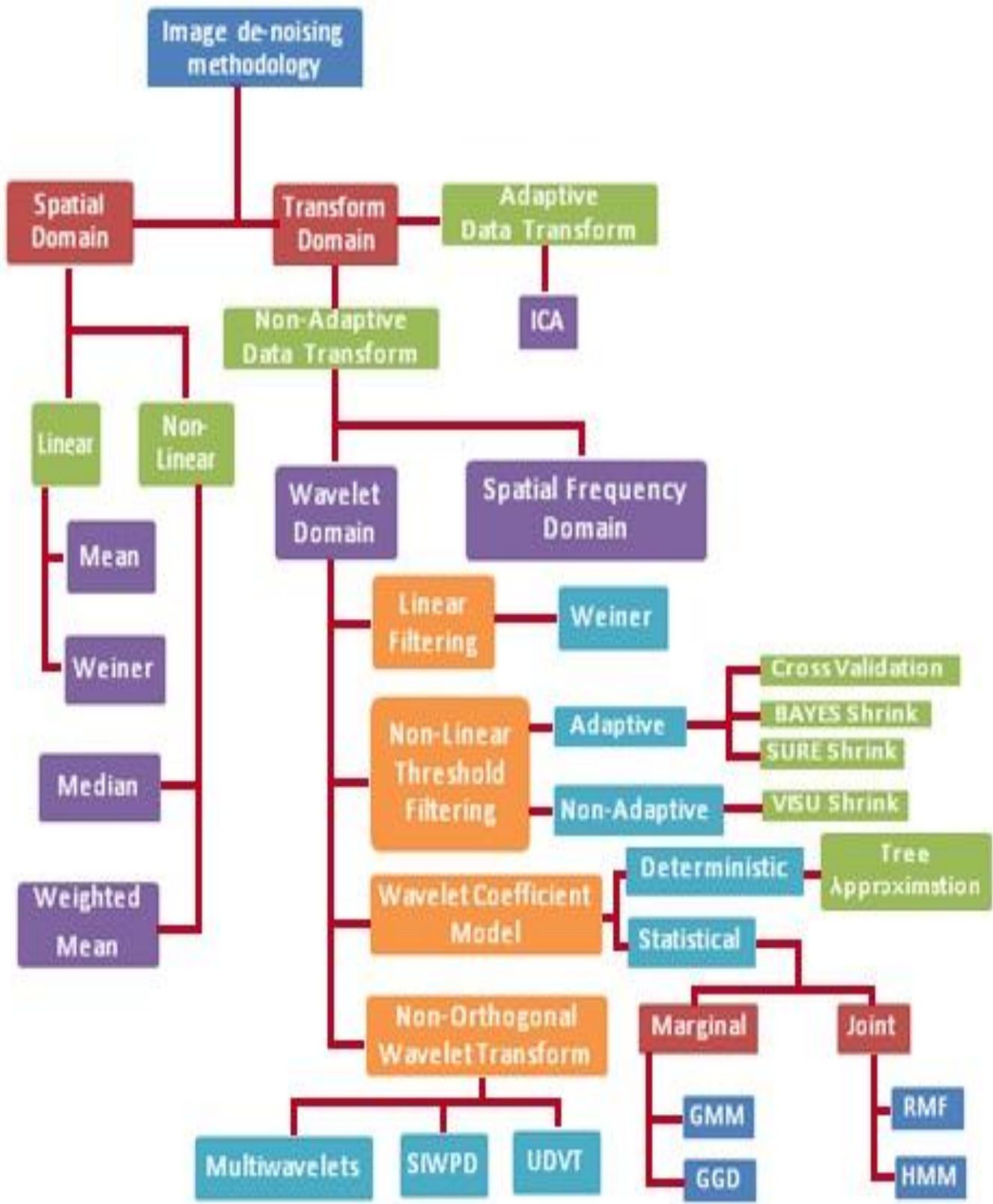

Fig. 9 Expanded classification of Image de-noising technique

*2.2.1 Non-Adaptive Thresholds*

VISU Shrink [12] is non-adaptive universal threshold, which depends only on number of data points. It has asymptotic equivalence suggesting best performance in terms of MSE when the number of pixels reaches infinity. VISU Shrink is known to yield overly smoothed images because its threshold choice can be unwarrantedly large due to its dependence on the number of pixels in the image.

*2.2.2 Adaptive Thresholds*

SURE Shrink [12] uses a hybrid of the universal threshold and the SURE [Stein's Unbiased Risk Estimator] threshold and performs better than VISU Shrink. Bayes Shrink [17], [18] minimizes the Bayes' Risk Estimator function assuming generalized Gaussian prior and thus yielding data adaptive threshold. Bayes Shrink outperforms SURE Shrink most of the times. Cross Validation [19] replaces wavelet coefficient with the weighted average of neighborhood coefficients to minimize generalized cross validation (GCV) function providing optimum threshold for every coefficient. The assumption that one can distinguish noise from the signal solely based on coefficient magnitudes is violated when noise levels are higher than signal magnitudes. Under this high noise circumstance, the spatial configuration of neighboring wavelet coefficients can play an important role in noise-signal classifications. Signals tend to form meaningful features (e.g. straight lines, curves), while noisy coefficients often scatter randomly.

*2.3 Non-orthogonal Wavelet Transforms*

Un-decimated Wavelet Transform (UDWT) has also been used for decomposing the signal to provide visually better solution. Since UDWT is shift invariant it avoids visual artifacts such as pseudo-Gibbs phenomenon. Though the improvement in results is much higher, use of UDWT adds a large overhead of computations thus making it less feasible. In [20] normal hard/soft thresholding was extended to Shift Invariant Discrete Wavelet Transform. In [21] Shift Invariant Wavelet Packet Decomposition (SIWPD) is exploited to obtain number of basic functions. Then using minimum description length principle the best basis function was found out which yielded smallest code length required for description of the given data. Then, thresholding was applied to denoise the data. In addition to UDWT, use of multi wavelets is explored which further enhances the performance but further increases the computation complexity. The multiwavelets are obtained by applying more than one function (scaling function) to given dataset. Multi wavelets possess properties such as short support, symmetry, and the most importantly higher order of vanishing moments. This combination of shift invariance and Multiwavelets is implemented in [22] which give superior results for the Lena image in context of MSE.

*2.4. Wavelet Coefficient Model*

This approach focuses on exploiting the multi resolution properties of Wavelet Transform. This technique identifies close correlation of signal at different resolutions by observing the signal across multiple resolutions. This method produces excellent output but is computationally much more complex and expensive. The modeling of the wavelet coefficients can either be deterministic or statistical.

*2.4.1 Deterministic*

The Deterministic method of modeling involves creating tree structure of wavelet coefficients with every level in the tree representing each scale of transformation and nodes representing the wavelet coefficients. This approach is adopted in [23]. The optimal tree approximation displays a hierarchical interpretation of wavelet decomposition. Wavelet coefficients of singularities have large wavelet coefficients that persist along the branches of tree. Thus if a wavelet coefficient has strong presence at particular node then in case of it being signal, its presence should be more pronounced at its parent nodes. If it is noisy coefficient, for instance spurious blip, then such consistent presence will be missing.

*2.4.2. Statistical Modeling of Wavelet Coefficients*

This approach focuses on some more interesting and appealing properties of the Wavelet Transform such as multi scale correlation between the wavelet coefficients local correlation between neighborhood coefficients etc. This approach has an inherent goal of perfect in the exact modeling of image data with use of Wavelet transform.

## IV. RESULT AND CONCLUSION

Performance of different denoising algorithms is measured by using quantitative performance measures such as peak signal-to-noise ratio (PSNR) and image enhancement factor (IEF).

Where MSE stands for mean square error, IEF stands for image enhancement factor, M x N is size of the image, Y represents the original image, $Y^\wedge$ denotes the de-noised image, and η represents the noisy image. The different denoising algorithms were applied to gray scale image of Lena containing noise density of 90%.

$$PSNR\ in\ dB = 10\log_{10}\left(\frac{255^2}{MSE}\right) \quad (1)$$

$$MSE = \frac{\sum_i \sum_j (Y(i,j) - \hat{Y}(i,j))^2}{M \times N} \quad (2)$$

$$IEF = \frac{\sum_i \sum_j (\eta(i,j) - Y(i,j))^2}{\sum_i \sum_j (\hat{Y}(i,j) - Y(i,j))^2} \quad (3)$$

TABLE I
COMPARISON OF IMAGE PARAMETERS FOR DIFFERENT IMAGE DENOISING FILTERS AT NOISE DENSITY OF 90%

| Filter | PSNR | IEF |
|---|---|---|
| MF | 6.5759 | 1.1712 |
| AMF | 8.0603 | 1.6499 |
| PSMF | 6.7847 | 1.2257 |
| DBA | 17.1205 | 13.2768 |
| MDBA | 17.2242 | 13.5976 |
| MDBUTMF | 17.9865 | 16.2066 |

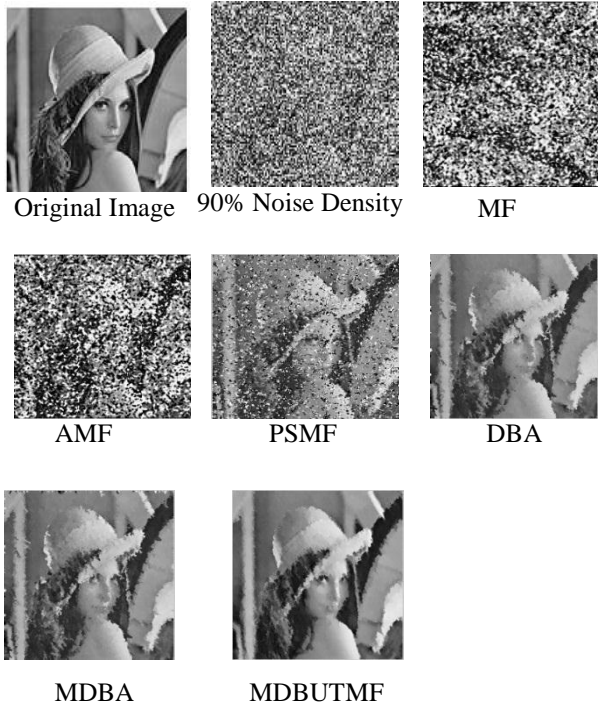

Original Image    90% Noise Density    MF

AMF    PSMF    DBA

MDBA    MDBUTMF

REFERENCES

[1] Castleman Kenneth R, *Digital Image Processing*, Prentice Hall, New Jersey, 1979.
[2] Reginald L. Lagendijk, Jan Biemond, *Iterative Identification and Restoration of Images*, Kulwer Academic, Boston, 1991.
[3] Scott E Umbaugh, *Computer Vision and Image Processing*, Prentice Hall PTR, New Jersey, 1998.
[4] Langis Gagnon, "Wavelet Filtering of Speckle Noise-Some Numerical Results," *Proceedings of the Conference Vision Interface* 1999, Trois Riveres.
[5] 1/$f$ noise, "Brownian Noise," http://classes.yale.edu/9900/math190a/OneOverF.html, 1999.
[6] Jacques Lévy Véhel, "Fraclab," www-rocq.inria.fr/fractales, May 2000.
[7] D. L. Donoho, "De-noising by soft-thresholding", IEEE Trans. Information Theory, vol.41, no.3, pp.613- 627, May1995. http://wwwstat.stanford.edu/~donoho/Reports/1992/denoisereleas e3.ps.Z
http://dx.doi.org/10.1109/18.382009
[8] R. Yang, L. Yin, M. Gabbouj, J. Astola, and Y. Neuvo, "Optimal weighted median filters understructural constraints," IEEE Trans. Signal Processing vol. 43, pp. 591–604, Mar. 1995.
http://dx.doi.org/10.1109/78.370615
[9] R. C. Hardie and K. E. Barner, "Rank conditioned rank selection filters for signal restoration," IEEE Trans. Image Processing, vol. 3, pp.192–206, Mar. 1994.
http://dx.doi.org/10.1109/83.277900
[10] A. Ben Hamza, P. Luque, J. Martinez, and R. Roman, "Removing noise and preserving details with relaxed median filters," J. Math. Imag. Vision, vol. 11, no. 2, pp. 161–177, Oct. 1999.
http://dx.doi.org/10.1023/A:1008395514426
[11] A.K.Jain,Fundamentals of digital image processing. Prentice-Hall, 1989.
[12] David L. Donoho and Iain M. Johnstone,"Ideal spatial adaption via wavelet shrinkage", Biometrika, vol.81, pp 425-455, September 1994.
http://dx.doi.org/10.1093/biomet/81.3.425
[13] David L. Donoho and Iain M. Johnstone., "Adapting to unknown smoothness via wavelet shrinkage", Journal of the American Statistical Association, vol.90, no432, pp.1200-1224, December 1995. National Laboratory, July 27, 2001.
[14] V. Strela. "Denoising via block Wiener filtering in wavelet domain". In 3rd European Congress of Mathematics, Barcelona, July 2000. Birkhäuser Verlag.
[15] H. Choi and R. G. Baraniuk, "Analysis of wavelet domain Wiener filters," in IEEE Int. Symp. Time- Frequency and Time-Scale Analysis, (Pittsburgh), Oct. 1998.
http://citeseer.ist.psu.edu/article/choi98analysis.html
[16] H. Zhang, Aria Nosratinia, and R. O. Wells, Jr., "Image denoising via wavelet-domain spatially adaptive FIR Wiener filtering", in IEEE Proc. Int. Conf. Acoustic., Speech, Signal Processing, Istanbul, Turkey, June 2000.
[17] E. P. Simoncelli and E. H. Adelson. Noise removal via Bayesian wavelet coring. In Third Int'l Conf on Image Proc, volume I, pages 379-382, Lausanne, September 1996. IEEE Signal Proc Society.
http://dx.doi.org/10.1109/ICIP.1996.559512
[18] H. A. Chipman, E. D. Kolaczyk, and R. E. McCulloch: 'Adaptive Bayesian wavelet shrinkage', J. Amer. Stat. Assoc., Vol. 92, No 440, Dec. 1997, pp. 1413-1421.
http://dx.doi.org/10.1080/01621459.1997.10473662
[19] Marteen Jansen, Ph. D. Thesis in "Wavelet thresholding and noise reduction" 2000.
[20] M. Lang, H. Guo, J.E. Odegard, and C.S. Burrus, "Nonlinear processing of a shift invariant DWT for noise reduction," SPIE, Mathematical Imaging: Wavelet Applications for Dual Use, April 1995.
[21] I. Cohen, S. Raz and D. Malah, Translation invariant denoising using the minimum description length criterion, Signal Processing, 75, 3, 201-223 (1999).
http://dx.doi.org/10.1016/S0165-1684(98)00234-5
[22] R. G. Baraniuk, "Optimal tree approximation with wavelets," in Proc. SPIE Tech. Conf. Wavelet Applications Signal Processing VII, vol. 3813, Denver, CO, 1999, pp. 196-207.